\documentclass[letterpaper, 10 pt, conference]{ieeeconf}
\IEEEoverridecommandlockouts    
\usepackage{subfiles}
\usepackage{graphics} 
\usepackage{epsfig} 
\usepackage{mathptmx} 
\usepackage{times} 

\usepackage{amsmath} 
\usepackage{amssymb}  
\usepackage{bbm}
\usepackage{xcolor}
\usepackage{graphicx}  
\usepackage{amsthm}
\usepackage{setspace}
\usepackage{soul}
\usepackage[ruled,vlined]{algorithm2e}
\usepackage{url}
\usepackage{cite} 
\usepackage{multirow}
\usepackage{hyperref}
\usepackage{cleveref} 
\usepackage{bm} 
\usepackage{array} 
\usepackage{booktabs} 
\usepackage{pgfplots}
\usepackage{tikz}
\usepackage{tikzscale}
\pgfplotsset{compat=1.3}
\pgfkeys{/pgf/number format/.cd,fixed}
\usetikzlibrary{plotmarks}
\usetikzlibrary{arrows.meta}
\usepgfplotslibrary{patchplots}

\usepackage{makecell} 

\newcommand{\skewcross}[1]{\lfloor #1 \rfloor^{\times}}
\makeatletter
\DeclareRobustCommand\frownotimes{\mathbin{\mathpalette\frown@otimes\relax}}
\newcommand{\frown@otimes}[2]{%
  \vbox{
    \ialign{##\cr
      \hidewidth$\m@th#1{}_\frown$\kern-\scriptspace\hidewidth\cr
      \noalign{\nointerlineskip\kern-1pt}
      $\m@th#1\otimes$\cr
    }%
  }%
}
\makeatother

\title{\LARGE \bf
Cerberus: Low-Drift Visual-Inertial-Leg Odometry \\ For Agile Locomotion
}

\author{Shuo Yang, Zixin Zhang, Zhengyu Fu, and Zachary Manchester
\thanks{Authors are with the Robotics Institute and the Department of Mechanical Engineering, Carnegie Mellon University, Pittsburgh, PA 15213 USA. Emails: \texttt{\{shuoyang, zixinz, zhengyuf, zmanches\}@andrew.cmu.edu} This work has been submitted to the IEEE for possible publication. Copyright may be transferred without notice, after which this version may no longer be accessible.}
}

\begin{document}
\maketitle
\thispagestyle{plain}
\pagestyle{plain}

\begin{abstract}
We present an open-source Visual-Inertial-Leg Odometry (VILO) state estimation solution, Cerberus, for legged robots that estimates position precisely on various terrains in real time using a set of standard sensors, including stereo cameras, IMU, joint encoders, and contact sensors. In addition to estimating robot states, we also perform online kinematic parameter calibration and contact outlier rejection to substantially reduce position drift.   
Hardware experiments in various indoor and outdoor environments validate that calibrating kinematic parameters within the Cerberus can reduce estimation drift to lower than 1\% during long distance high speed locomotion. Our drift results are better than any other state estimation method using the same set of sensors reported in the literature. Moreover, our state estimator performs well even when the robot is experiencing large impacts and camera occlusion. The implementation of the state estimator, along with the datasets used to compute our results, are available at \url{https://github.com/ShuoYangRobotics/Cerberus}.

\end{abstract}

\section{Introduction}\label{sec:intro}

Using onboard sensors to estimate a robot's state (typically body pose and velocity) is a critical functionality for legged robots \cite{bloesch2013state, bledt2018cheetah, camurri2020pronto, wisth2022vilens}. A sensor solution including only one pair of stereo cameras and critical proprioceptive sensors (IMU, joint encoders, and foot contact sensors) serves as an ideal choice for resource-constrained robots because this set of sensors is low cost, compact, and has low power consumption \cite{bloesch2017state}. We call a state estimator using this sensing solution a Visual-Inertial-Leg Odometry (VILO) estimator. VILO fuses data from different sensors by constructing observation models that predict measurements given robot states. Observation models combined with a dynamics model of the robot form a factor graph \cite{dellaert2017factor} describing a nonlinear optimization problem whose solution is the maximum-likelihood state estimate. Prior work \cite{hartley2018hybrid, wisth2019robust, kim2022step} has shown that VILO outperforms methods that only utilize a subset of the aforementioned sensors, such as Visual-Inertial-Odometry (VIO) \cite{li2013high} or Leg Odometry (LO) \cite{camurri2020pronto} alone. 

A key feature of VIO estimators is online calibration of IMU biases using visual measurements\cite{forster2015imu}. Other key error sources in VIO have recently been systematically addressed \cite{qin2018online}. However, in the VILO setting, systematic error analysis has yet to be established for leg sensors (joint encoders and contact sensors). Prior work has identified that when generating body velocity estimates using LO, error sources such as foot slippages, impacts, rolling contacts, and kinematic parameter errors \cite{bloesch2013slip, wisth2020preintegrated, yang2022online} could degrade velocity estimation accuracy. However, no prior work has studied how to handle these error sources in a VILO estimator. 

Since different legged robots have different leg configurations, locomotion strategies, and sensor qualities, it is hard to fairly compare the performance of different VILO implementations. An open-source baseline VILO implementation and public datasets are needed for the benefit of the entire legged robot community.

\begin{figure}
    \centering
   \begin{minipage}[b]{.35\linewidth}
    \resizebox{1.0\linewidth}{!}{%
      \includegraphics[width=\linewidth]{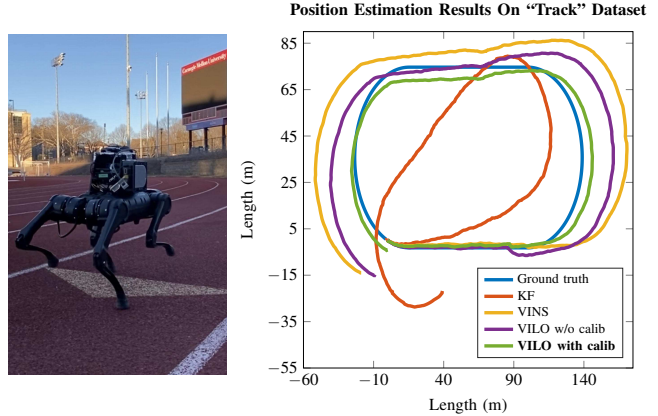}
      }
      \vspace{5pt}
   \end{minipage}%
   \hfill
   \begin{minipage}[b]{.65\linewidth}
    \resizebox{1.0\linewidth}{!}{%
        \input{images/title_pic}
    }
   \end{minipage}
    
    \caption{
        On the A1 robot, the Cerberus algorithm has lower than 1\% position estimation drift after traveling 450m on standard stadium track, 
        better than any baseline methods and better than any drift performance reported in literature using the same set of sensors. The ground truth is obtained using dimensions of standard running track.
      }
    \label{fig:title_pic}
\end{figure}

As a first step toward establishing a standard VILO benchmark, we present a state-of-the-art real-time VILO algorithm called Cerberus that incorporates kinematic calibration for improved accuracy, as well as several datasets from two different quadruped robots. The algorithm implementation uses standard ROS interfaces to process sensor data and publish estimation results, and the datasets are in the format of ROS bags \cite{ros}. Docker \cite{merkel2014docker} provides easy installation of a unified testing environment. 
Our contributions are:
\begin{itemize}
    \item Cerberus, a VILO algorithm that estimates kinematic parameters online to achieve drift rates lower than any other results reported in the literature.
    \item Datasets collected on multiple robots in various indoor and outdoor environments to benchmark the Cerberus implementations. 
    \item Open-source algorithm implementations using standard ROS interfaces that can be readily adapted to different robots and sensor configurations.
\end{itemize}

This paper is organized as follows. In Section \ref{sec:related} we review related work. Section \ref{sec:background} introduces notation and provides background. Section \ref{sec:vilo} presents a basic VILO algorithm. Section \ref{sec:technical} derives an online kinematic calibration method in the Cerberus. Section \ref{sec:experiments} describes details of the algorithm implementation and presents hardware experiment results. Section \ref{sec:conclude} summarizes our conclusions.

\section{Related Work}\label{sec:related}

Using multiple sensors to estimate the physical state of a robot is one of the central topics of robotics. Although the Global Position System (GPS) can provide a good position estimation solution, many robots need to operate in GPS-denied environments. Visual odometry (VO) \cite{scaramuzza2011visual}, which estimates robot pose using a monocular or a stereo camera, can provide a solution in these settings. By matching features across image sequences, feature locations constrain the possible motion of the camera so displacement can be solved from multiple-view geometry \cite{hartley2003multiple}. To improve the robustness and accuracy of the estimation, VIO \cite{li2013high} uses both the camera and the IMU as motion constraints. Preintegration \cite{forster2015imu}, and factor graphs and their associated Maximum a posteriori (MAP) estimation algorithms \cite{dellaert2017factor} can also help VIO to exploit problem structure, hence reducing computation cost. After the development of several VIO algorithms \cite{li2013high, sun2018robust, qin2018vins}, researchers continue to study how to reject different error sources in VIO including IMU biases, sensor time delay, and extrinsic parameter errors \cite{qin2018online}. The position drift percentage, measuring how many meters the estimation deviates from the ground truth after traveling 100 meters, is often used as an important performance metric. Once the error sources are properly addressed, position drift of a VIO estimator can be as low as 0.29\% on drones \cite{qin2018vins}.

Early legged robot state estimation work focused on fusing IMU and LO data using a Kalman Filter
(KF), and analyzed error sources in this setting. A legged robot often experiences link deformations, foot slippage, and excessive body rotation due to repeated impacts with the ground, all of which may lead to incorrect or biased velocity estimation. \cite{bloesch2013state} showed that using body IMU, joint encoders, and foot contact sensors can recover robot pose, velocity and IMU biases. A similar linear KF formulation is proposed in \cite{bledt2018cheetah}. The invariant EKF is proposed in \cite{hartley2020contact} to improve orientation estimation convergence. The non-slipping assumption of LO relies on accurate contact sensing \cite{camurri2017probabilistic} or slipping rejection mechanisms \cite{bloesch2013slip}. Some algorithms estimate contacts using kinematic information \cite{hwangbo2016probabilistic}, eliminating the dependency on foot contact sensors. \cite{yang2022online} identifies forward kinematic parameter errors due to link length changes and rolling contacts as another major error source in LO.  

The factor graph formulation used in VIO can be easily extended to include the LO motion constraint, which leads to the VILO estimator \cite{hartley2018hybrid, wisth2019robust, kim2021legged}. \cite{wisth2019robust} uses the velocity estimation result of a KF as the motion constraint. Contact preintegration is developed in \cite{hartley2018hybrid}, but bias correction is not performed. \cite{wisth2020preintegrated} describes the LO velocity bias and models it as a linear term that can be corrected in the preintegration. However, this bias model does not explain the source of the bias and its physical meaning. With the velocity bias model, \cite{wisth2022vilens} further shows that VILO can reach around 1\% position drift with the aid of lidar, though their VILO implementation and datasets are not publicly available. 

\section{Background}\label{sec:background}

We now introduce relevant notations and review some concepts from legged robot state estimation that are previously used in \cite{yang2022online}. In general, we use lowercase letters for scalars and frame abbreviations, boldface lowercase letters for vectors, and upper case letters for matrices and vector sets. The operation $[a;b;c]$ vertically concatenates elements $a$, $b$ and $c$. 
The operator $\skewcross{\bm{v}}$ converts a vector $\bm{v} = [v_1;v_2;v_3]\in \mathbb{R}^3$ into the skew-symmetric ``cross-product matrix,''
\begin{equation}
\skewcross{\bm{v}} = 
\begin{bmatrix}
  0 & -v_3 & v_2 \\
v_3 &    0 & -v_1 \\
-v_2 & v_1 & 0  
\end{bmatrix} ,
\end{equation}
such that $\bm{v} \times \bm{x} = \skewcross{\bm{v}} \bm{x}$. Lastly, $\hat{\bm{a}}$ indicates an estimate of $\bm{a}$.

\subsection{Coordinate Frames \& Quaternions}
\begin{figure}
    \centering
    \includegraphics[width=0.8\linewidth]{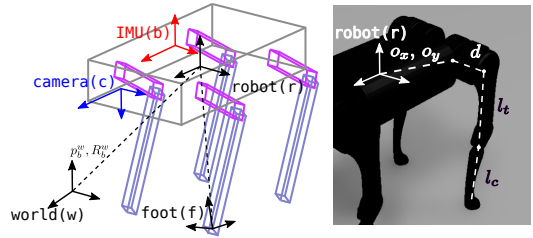}
    \caption{Frames \& Kinematic parameters of A1 robot.}
    \label{fig:frame}
\end{figure}
Important coordinate frames are shown in Fig. \ref{fig:frame}. For simplicity, we assume that the IMU frame and the robot's body frame coincide. We use $\bm{p}$ and $\bm{q}$ to denote the translation vector and the unit-quaternion rotation, respectively, from the robot body frame to the world frame. We follow the quaternion convention defined in \cite{jackson2021planning}. A quaternion $\bm{q} = [q_w; \bm{q}_v]$ has a scalar part $q_w$ and a vector part $\bm{q}_v = [q_x; q_y; q_z] \in \mathbb{R}^3$. We define the two matrices, 
$$
L(\bm{q}) = 
\begin{bmatrix}
  q_s & -\bm{q}_v^{\top} \\
  \bm{q}_v & q_sI+\skewcross{\bm{q}_v}
\end{bmatrix}
\ \ \text{and} \ \
R(\bm{q}) = 
\begin{bmatrix}
  q_s & -\bm{q}_v^{\top} \\
  \bm{q}_v & q_sI-\skewcross{\bm{q}_v}
\end{bmatrix},
$$
such that the product of two quaternions can be written as,
\begin{equation}
\bm{q}_1 \otimes \bm{q}_2 = L(\bm{q}_1)\bm{q}_2 = R(\bm{q}_2)\bm{q}_1.
\end{equation}
It can also be shown that the inverse of a unit quaternion $\bm{q}$ is $\bm{q}^{-1} = [q_w;-\bm{q}_v]$ and $\bm{q}\otimes\bm{q}^{-1}=\bm{q}_I = [1;\bm{0}]$, the identity quaternion.
We also introduce a matrix $B = \begin{bmatrix}
  0 \\I_{3x3}
\end{bmatrix}$ that converts a vector in $\mathbb{R}^3$ to a quaternion with zero scalar part. The rotation matrix $A(\bm{q})$ can then be written in terms of $\bm{q}$ as,
\begin{equation}
    A(\bm{q}) = B^{\top}L(\bm{q})R(\bm{q})^{\top}B.
\end{equation}

Small rotation approximations play an important role in orientation estimation. We parameterize small rotations using Rodrigues parameters $\delta \bm{\theta}\in \mathbb{R}^3$ and map them into unit quaternions using the Cayley map \cite{jackson2021planning}:
\begin{equation}
\delta \bm{q} = \Phi(\delta \bm{\theta}) = \frac{1}{\sqrt{1+\|\delta \bm{\theta}\|^2}}\begin{bmatrix}
  1 \\ \delta \bm{\theta}
\end{bmatrix}.
\end{equation}
Assuming the true orientation of a robot is $\bm{q}$ and our estimate is $\hat{\bm{q}}$, we define the error as $\delta \bm{q} = \hat{\bm{q}}^{-1}\otimes\bm{q}$. We use the inverse Cayley map \cite{jackson2021planning}
$
\Phi^{-1}(\bm{q}) = \bm{q}_v/q_s
$
to convert the estimation error into Rodrigues parameters $\delta \bm{\theta}$. Therefore
$
    \bm{q} = L(\hat{\bm{q}})\Phi(\delta \bm{\theta}).
$

Where necessary, we use superscripts and subscripts to explicitly indicate the frames associated with rotation matrices and vectors, so $A^{a}_{b}\cdot p$ means the matrix transforms a vector $p$ represented in coordinate frame $b$ into coordinate frame $a$ \cite{murray2017mathematical}. If frame $b$ is time varying, $A^{a}_{b_k}$ indicates the frame at time $k$. When the context is estimating the robot body frame $b_k$ in the world frame $w$, we would also write $A_k$ instead of $A^{w}_{b_k}$ for brevity. Similarly, $p^{w}_{b_k}$ or $p_{k}$ defines the origin vector of frame $b_k$ in the world.

\begin{figure}
  \centering
  \includegraphics[width=\linewidth]{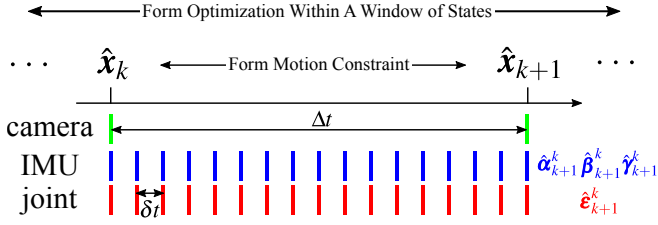}
  \caption{Illustration of Preintergation.}
  \label{fig:integration}
\end{figure}
\subsection{Forward Kinematics \& Leg Odometry Velocity}\label{sec:fk-lo}
In this section we review forward kinematics and describe how to infer body velocity. We define $\bm{\phi}$ as a vector containing all joint angles of the robot's $j$'th leg, and $\dot{\bm{\phi}}$ the corresponding joint angle velocities. The forward kinematics function is denoted as
$
\bm{p}_{f} = g(\bm{\phi}, \bm{\rho}) \in \mathbb{R}^3
$, whose output is the foot position in the robot body frame. $\bm{\rho}$ is a set of kinematic parameters of interest, such as link lengths and motor offsets \cite{yang2022online}. The derivative of this equation with respect to $\bm{\phi}$ leads to the Jacobian matrix $J(\bm{\phi}, \bm{\rho})$ that maps $\dot{\bm{\phi}}$ into the foot's linear velocity in the body frame:
\begin{equation}
\bm{v}_{f} = \dot{\bm{p}}_{f} = J(\bm{\phi}, \bm{\rho})\dot{\bm{\phi}}.
\end{equation}

Assuming the $j$'th foot is in contact with the ground and does not slip, $g$ and $J$ can be used to calculate the body velocity of the robot. Let $\bm{p}^w_f$ denote the foot position in the world frame (see Fig. \ref{fig:frame}); It is a function of the robot's body position $\bm{p}$ and joint angles $\bm{\phi}$:

\begin{equation}\label{eqn::foot_position}
\bm{p}^w_f = \bm{p} + A(\bm{q})\bm{p_{f}} = \bm{p} + A(\bm{q})g(\bm{\phi}, \bm{\rho}).
\end{equation}
Let the time derivative of $\bm{p}^w_f$ be $\bm{v}^w_f$. The no-slip assumption means $\bm{v}^w_f = 0$. Therefore, by differentiating \eqref{eqn::foot_position}, we have 
\begin{equation}\label{eqn::foot_position_derive}
  0 = \bm{v}^w_f = \bm{\dot{p}}^w_f = \bm{\dot{p}} + A(\bm{q})\frac{d}{dt}g(\bm{\phi}, \bm{\rho}) + \frac{d}{dt}A(\bm{q})g(\bm{\phi}, \bm{\rho}).
\end{equation} 
It is shown in \cite{murray2017mathematical} that $\frac{d}{dt}A(\bm{q}) = A(\bm{q})\skewcross{\bm{\omega}}$, where $\bm{\omega}$ is the robot body angular velocity. We define $\bm{v} = \bm{\dot{p}}$, then 
from \eqref{eqn::foot_position_derive} we derive an expression for the body velocity in the world frame: 
\begin{equation}
\label{eqn:lo}
\bm{v} = -A(\bm{q})[ J(\bm{\phi}, \bm{\rho})\dot{\bm{\phi}} + \skewcross{\bm{\omega}}g(\bm{\phi}, \bm{\rho})].
\end{equation}
This velocity is called the LO velocity because its integration is the body displacement\cite{lin2005leg}. During legged locomotion, the kinematic parameters $\bm{\rho}$, which conventionally are deemed constant, change due to link deformations and rolling contacts \cite{yang2022online}. Therefore, the parameter error can be viewed as a ``bias'' of the LO velocity measurement.

\section{Visual-Inertial-Leg Odometry}\label{sec:vilo}

A typical VILO framework \cite{wisth2020preintegrated, hartley2018hybrid, kim2022step} keeps track of the estimation of a list of past N states $\hat{\bm{x}}_k$ and M camera feature locations $\hat{\lambda}_l$ as $\mathcal{X} = \{\hat{\bm{x}}_0, \hat{\bm{x}}_1, \dots \hat{\bm{x}}_N, \hat{\lambda}_0, \hat{\lambda}_1, \dots \hat{\lambda}_M\}$. The robot state is  $\hat{\bm{x}}_k = [\hat{\bm{p}}_k; \hat{\bm{q}}_k; \hat{\bm{v}}_k; \hat{\bm{b}}_{ak}; \hat{\bm{b}}_{\omega k}]$, where $\hat{\bm{p}}_k \in \mathbb{R}^3$ is the robot position in the world frame, $\hat{\bm{q}}_k$ is the robot's orientation quaternion, and $\hat{\bm{v}}_k \in \mathbb{R}^3$ is the linear velocity of the robot's body represented in the world frame. $\hat{\bm{b}}_{ak} \in \mathbb{R}^3$ and $\hat{\bm{b}}_{\omega k} \in \mathbb{R}^3$ are IMU accelerometer bias and gyroscope bias. 
A new state $\hat{\bm{x}}_k$ is created each time $t_k$ when a new camera image arrives. Also, sensors on the robot generate measurements $Z_t=\{\hat{\bm{a}}_{m}(t),\hat{\bm{\omega}}_{m}(t),\hat{\bm{\phi}}_{j}(t),\hat{\dot{\bm{\phi}}}_{j}(t)\}$ and $\Lambda_{t}$ periodically, where $\hat{\bm{a}}_{m}$ and $\hat{\bm{\omega}}_{m}$ are IMU linear acceleration and angular velocity, $\hat{\bm{\phi}}_{j}$ and $\hat{\dot{\bm{\phi}}}_{j}$ are joint angle and joint angle velocity for each leg $j$, and $\Lambda_{t}$ is a set of feature coordinates on the camera images who have known associations with feature locations in $\mathcal{X}$. We denote $\mathcal{Z}$ as all measurements between state $\hat{\bm{x}}_0$ and $\hat{\bm{x}}_N$. We also denote subsets $\mathcal{X}_{sub} \subset \mathcal{X}$ and $\mathcal{Z}_{sub} \subset \mathcal{Z}$. The VILO constructs a nonlinear least-squares problem to find $\mathcal{X}$ as the solution of 

\begin{equation}\label{eqn:map}
    \min_{\mathcal{X}^*} \bigg\{ \sum_i \bigg\| \bm{r}_i(\mathcal{X}_{sub}, \mathcal{Z}_{sub})\bigg\|^2_{P_i} \bigg\},
\end{equation}
where each term $\bm{r}_i(\mathcal{X}_{sub}, \mathcal{Z}_{sub})$ defines a measurement residual function. Ideally the cost should be \textbf{0} at optimal solution $\mathcal{X}^*$. $P_i$ is a weighting matrix that encodes the relative uncertainty in each $\bm{r}_i$, and also takes the same set of inputs. Problem \eqref{eqn:map} can be solved by nonlinear optimization methods \cite{dellaert2017factor}. The core technical challenge is to design cost functions and their uncertainties leveraging all available sensor data. Additionally, a VILO estimator usually has other mechanisms to ensure real-time computation, such as visual feature tracking and marginalization. See \cite{qin2018vins, wisth2019robust} for more details.

\subsection{Preintegration}
A key technique used in VIO and VILO to improve computation efficiency is preintergration. When fusing camera data and IMU data with different frequencies, preintegration \cite{forster2015imu} is used to integrate multiple IMU measurements between two camera image times into a single ``motion constraint'' in the cost function, so the estimator only needs to add states at the camera frequency instead of keeping up with the much higher frequency of the IMU. More importantly, it is well known that IMUs are biased \cite{adams1956introduction}, and biases should be estimated along with robot physical states. When the estimator updates IMU biases, IMU preintegration can avoid integrating measurements again by directly updating the integration term using its first order approximation. IMU preintergration is used in several real-time VIO algorithms \cite{usenko2016direct, sun2018robust, qin2018vins}. Similarly, contact preintegration is used to integrate joint encoder data into motion constraints in VILO \cite{hartley2018legged}. 

We assume there are $L$ IMU measurements between state $\hat{\bm{x}}_k$ and $\hat{\bm{x}}_{k+1}$, and that each IMU measurement arrives $\delta t$ after the previous one. Let $i \in \{1 \dots L\}$ be the measurement index and $\Delta t = t_{k+1} - t_{k}$, then $t_{1} = t_k$ and $t_L = t_{k+1}$. As shown in Figure \ref{fig:integration}, we can integrate these IMU measurements into a single motion measurement.

First, let $\hat{\bm{\gamma}}^{k}_{i}$ denote quaternion rotation from frame $b_k$ to frame $b_i$, the robot body frame at time $t_i$. Starting from $\hat{\bm{\gamma}}^{k}_{k} = \bm{q}_I$, we can calculate 
\begin{align}
  \hat{\bm{\gamma}}^{k}_{i+1} & = R(\frac{1}{2}\begin{bmatrix}0 \\ (\hat{\bm{\omega}}_{m}(t_i)-\hat{\bm{b}}_{\omega k})\delta t\end{bmatrix})  \hat{\bm{\gamma}}^{k}_i \label{eqn:pre-int-terms-gamma},
\end{align}
which recursively leads to $\hat{\bm{\gamma}}^{k}_{k+1}$, a measurement of the rotation difference between $\hat{\bm{q}}_k$ and $\hat{\bm{q}}_{k+1}$. Another two recursive relations can be derived using acceleration data as
\begin{align}
  \hat{\bm{\alpha}}^{k}_{i+1} & =  \hat{\bm{\alpha}}^{k}_{i} + \hat{\bm{\beta}}^{k}_{i} \delta t ,\ \text{and}\\
  \hat{\bm{\beta}}^{k}_{i+1} & = \hat{\bm{\beta}}^{k}_{i}+ A(\hat{\bm{\gamma}}^{k}_i) (\hat{\bm{a}}_{m}(t_i)-\hat{\bm{b}}_{ak}) \delta t\label{eqn:pre-int-terms-alpha} , 
\end{align}
such that $\hat{\bm{\alpha}}^{k}_{k+1}$ and $\hat{\bm{\beta}}^{k}_{k+1}$ measure position and velocity differences between two states. These so called preintegration terms \cite{forster2015imu} describe a cost function on states as \cite{qin2018vins}
\begin{align}
  & \bm{r}(\hat{\bm{x}}_k, \hat{\bm{x}}_{k+1},Z_{\Delta k}) =  \nonumber\\
  & \begin{bmatrix}
      A(\hat{\bm{q}}_k)^T(\hat{\bm{p}}_{k+1} - \hat{\bm{p}}_{k} + \frac{1}{2}g^w\Delta t^2 - \hat{\bm{v}}_{k}\Delta t) - \hat{\bm{\alpha}}^{k}_{k+1}\\
      \Phi^{-1}(\hat{\bm{q}}_k^{-1} \otimes \hat{\bm{q}}_{k+1} \otimes (\hat{\bm{\gamma}}^{k}_{k+1})^{-1}) \\
      A(\hat{\bm{q}}_k)^T(\hat{\bm{v}}_{k+1} + g^w\Delta t - \hat{\bm{v}}_{k}) - \hat{\bm{\beta}}^{k}_{k+1} \\
      \hat{\bm{b}}_{ak+1} - \hat{\bm{b}}_{ak} \\
      \hat{\bm{b}}_{\omega k+1} - \hat{\bm{b}}_{\omega k} \\
  \end{bmatrix}, \label{eqn:residual}
\end{align}
where $Z_{\Delta k}$ represents all measurements during $\Delta t$. 

The error dynamics \cite{qin2018vins} of $\bm{r}$ as 
\begin{align}
    & {\bm{e}}_{i+1}  =  
  \begingroup 
    \setlength\arraycolsep{1pt}
    \begin{bmatrix}
  I & I\delta t & 0 & 0 & 0 \\
  0 & I & -A(\hat{\bm{\gamma}}^{k}_i)\skewcross{\hat{\bm{a}}_{m}(t_i)-\hat{\bm{b}}_{ak}}\delta t & -A(\hat{\bm{\gamma}}^{k}_i)\delta t & 0 \\
  0 & 0 & I-\skewcross{\hat{\bm{\omega}}_{m}(t_i)-\hat{\bm{b}}_{\omega k}}\delta t & 0 & -I\delta t \\
  0 & 0 & 0 & I & 0 \\
  0 & 0 & 0 & 0 & I 
  \end{bmatrix}
  \endgroup
  \bm{e}_i \nonumber \\
   + & 
   \begingroup 
     \setlength\arraycolsep{1pt}
  \begin{bmatrix}
    0 & 0 & 0 & 0 \\
    -A(\hat{\bm{\gamma}}^{b_k}_i)\delta t & 0 & 0 & 0 \\
    0 & -I\delta t & 0 & 0 \\
    0 & 0 & I\delta t & 0 \\
    0 & 0 & 0 & I\delta t \\
  \end{bmatrix} 
  \endgroup
  \begin{bmatrix}\bm{n}_a \\ \bm{n}_{\omega} \\ \bm{n}_{ba}\\ \bm{n}_{b\omega }\end{bmatrix}= F_i \bm{e}_i + G_i \bm{n}_{IMU}, \label{eqn:error-dynamics}
\end{align}
where $\bm{n}_{a}$ and $\bm{n}_{\omega}$ are IMU sensor measurement noises and $\bm{n}_{ba}$ and $\bm{n}_{b\omega}$ are random walk noises for IMU biases. $\bm{e}_i = [\delta \bm{\alpha}^{k}_i; \delta \bm{\beta}^{k}_i; \delta \bm{\theta}^{k}_i; \delta \bm{b}_{ai}; \delta \bm{b}_{\omega i}]$ is a vector describing the errors between preintegration terms and their ``true'' values after each IMU measurement integration \cite{qin2018vins}. $\delta \bm{\alpha}^{k}_{i} = \bm{\alpha}^{k}_{i} - \hat{\bm{\alpha}}^{k}_{i}$, $\delta \bm{\beta}^{k}_{i}$, and $\bm{\gamma}^{k}_{i} = L(\hat{\bm{\gamma}}^{k}_{i})\Phi(\delta \bm{\theta}^{k}_i)$. Details of the derivation can be seen in \cite{qin2018vins}.

Let $Q$ be the noise covairance matrix of $\bm{n}_{IMU}$. We can also recursively calculate $P_{k+1}^k$ and $J_{k+1}$, the error jacobian, as follows
\begin{align}
  P^k_{i+1} = F_iP^k_{i}F_i^T +  G_iQG_i^T, P^k_{1} = 0, \label{eqn:p} \\
  J_{i+1} = F_iJ_{i}, J_{i} = I. \label{eqn:j}
\end{align}

The error jacobian can greatly reduce VILO computation time: When solving Problem \eqref{eqn:map} using numerical methods, a solver iteratively calculates state update vectors $\delta \bm{x}$, and the update will change IMU biases. Instead of reintegrating the preintegration terms that depend on IMU biases, with the error jacobian, we can directly update the preintegration terms, for example, as 
\begin{equation}
  \bm{\alpha}^{k}_{k+1} = \hat{\bm{\alpha}}^{k}_{k+1} + J^{\alpha}_{a}\delta \bm{b}_{a} + J^{\alpha}_{\omega }\delta \bm{b}_{\omega }
\end{equation}
to get their revised values, where $J^{\alpha}_{a}$ and $J^{\alpha}_{\omega }$ are blocks in $J_{k+1}$ that correspond to $\partial \bm{\alpha}/\partial \bm{b}_{a}$ and $\partial \bm{\alpha}/\partial \bm{b}_{\omega }$.

\section{Kinematic Calibration In Preintegration}\label{sec:technical}

\begin{table}[]
    \centering
    \small{
    \begin{tabular}{|c|c|c|c|}
    \hline
      Type \& Model & No. & Freq. & Output Description  \\
    \hline
     D435 camera \cite{realsense}  & 1   &    15Hz   & A pair of stereo images \\
    \hline
      Robot built-in IMU  & 1 &   500Hz   & \makecell{Linear acceleration \& \\ angular velocity} \\
    \hline
      \makecell{Robot built-in \\joint encoder}& 12  &   500Hz  & \makecell{Joint motor angles \& \\ angle velocities} \\
    \hline
     \makecell{Robot built-in \\contact sensor} & 4  &   500Hz  & Binary foot contact flag\\
    \hline
    \end{tabular}
    }
    \caption{VILO Sensor List}
    \label{tab:hardware}
\end{table}
In this section we show, in the Cerberus, how to estimate $\bm{\rho}$ for each leg discussed in Section \ref{sec:fk-lo} by including them into the state so $\hat{\bm{x}}_k = [\hat{\bm{p}}_k; \hat{\bm{q}}_k; \hat{\bm{v}}_k; \hat{\bm{b}}_{ak}; \hat{\bm{b}}_{\omega k}; \hat{\bm{\rho}}_{jk}]$, where $j$ is the leg index. For brevity, we only describe the case $j=1$ but the algorithm can easily apply to robots with more legs. 

\subsection{Contact Preintegration}
For a leg that has non-slipping contact with the ground, \eqref{eqn:lo} describes body velocity estimation through LO. This velocity can be integrated into a body displacement. We again focus on integrating measurements between state $\hat{\bm{x}}_k$ and $\hat{\bm{x}}_{k+1}$ including sensor data from leg sensors, then have a revised constraint equation
\begin{align}
    & \bm{r}'(\hat{\bm{x}}_k, \hat{\bm{x}}_{k+1},Z_{\Delta k}) = \begin{bmatrix}
        \bm{r}(\hat{\bm{x}}_k, \hat{\bm{x}}_{k+1},Z_{\Delta k}) \\
        A(\hat{\bm{q}}_k)^T(\hat{\bm{p}}_{k+1} - \hat{\bm{p}}_{k}) - \hat{\bm{\epsilon}}^{k}_{k+1} \\
        \hat{\bm{\rho}}_{k+1} - \hat{\bm{\rho}}_{k} \\
    \end{bmatrix}, \label{eqn:residual-leg}
  \end{align}
where $\hat{\bm{\epsilon}}^{k}_{k+1}$ is the integration result of
\begin{align}
    \hat{\bm{\epsilon}}^{k}_{i+1} & = \hat{\bm{\epsilon}}^{k}_{i} + A(\hat{\bm{\gamma}}^{b_k}_i)\hat{\bm{v}}_{i}\delta t,\ \text{where} \label{eqn:pre-int-terms-epsilon}
\end{align}
\begin{equation}\label{eqn:hat_vm}
    \hat{\bm{v}}_{i} = -[J(\hat{\bm{\phi}},\hat{\bm{\rho}})\hat{\dot{\bm{\phi}}} + \skewcross{\hat{\bm{\omega}}-\hat{\textbf{b}}_{\omega k}} g(\hat{\bm{\phi}},\hat{\bm{\rho}})].
\end{equation} 

Comparing to \eqref{eqn:residual}, \eqref{eqn:residual-leg} introduces the LO velocity integration as a measurement model of body positions. The term $\hat{\bm{\epsilon}}^{k}_{k+1}$ depends on sensor measurements, $\hat{\bm{b}}_{\omega k}$, and $\hat{\bm{\rho}}_{k}$. A version without kinematic parameter dependency is previously derived in \cite{hartley2018hybrid}. The error of this measurement, defined as $\bm{e}'_i = [\bm{e}_i; \delta \bm{\epsilon}^k_{i}; \delta \bm{\rho}_i]$, has dynamics
\begin{align}
    \bm{e}'_{i+1} & = 
    \begingroup 
    \setlength\arraycolsep{2pt}
    \begin{bmatrix}
        & & F_i & & \bm{0} & \\
        0 & I & -A(\hat{\bm{\gamma}}^{b_k}_i)\skewcross{\hat{\bm{v}}_{i}}\delta t  & 0 &\bm{\zeta}\delta t & 0 & \bm{\kappa}\delta t   \\
        0 & 0 & 0  & 0 & 0 & 0 & 0   
    \end{bmatrix}
    \endgroup
    \bm{e}'_i \nonumber \\
    + & 
    \begingroup 
    \setlength\arraycolsep{2pt}\begin{bmatrix}
         & & G_i &  &   & & \bm{0} \\
        0 & \bm{\zeta}\delta t & 0 & 0 & \bm{\eta}\delta t & A(\hat{\bm{\gamma}}^{b_k}_i)J\delta t & I\delta t & 0 \\
        0 & 0 & 0 & 0 & 0 & 0 & 0   & I\delta t 
    \end{bmatrix}
    \endgroup
    \begin{bmatrix}
        \bm{n}_{IMU} \\
        \bm{n}_{\phi} \\
        \bm{n}_{\dot{\phi}} \\
        \bm{n}_{v} \\
        \bm{n}_{\rho}
    \end{bmatrix}, \label{eqn:error-dyn-leg}
\end{align}
in which $J$ is short for $J(\bm{\phi},\hat{\bm{\rho}})$, the forward kinematics Jacobian. $\bm{e}_t$, $\bm{n}_t$, $F_t$, and $G_t$ are defined in \ref{eqn:error-dynamics}. The definitions of $\bm{\zeta}$, $\bm{\eta}$, and $\bm{\kappa}$, along with the derivation of the error dynamics, are in the Appendix. $\bm{n}_{\phi} \sim \mathcal{N}(0,\,\sigma_\phi^{2})$ and $\bm{n}_{\dot{\phi}} \sim \mathcal{N}(0,\,\sigma_{\dot{\phi}}^{2})$ are the measurement noise of joint angle and joint angle velocity. $\bm{n}_\rho \sim \mathcal{N}(0,\,\sigma_\rho^{2})$ is the kinematic parameter random walk noise. $\bm{n}_{v} \sim \mathcal{N}(0,\,\sigma_{v}^{2})$ is the uncertainty of the contact preintegration motion constraint.


From the error dynamics, we can get $P_{k+1}^k$ and $J_{k+1}$ as in \eqref{eqn:p} and \eqref{eqn:j}. Then Jacobians such as $J^{\epsilon}_{\rho} = \frac{\partial \epsilon^{k}_{k+1}}{\partial \rho}$ extracted from $J^k_{k+1}$ can allow fast preintegration updates:
\begin{align}
    \bm{\epsilon}^{k}_{k+1} & = \hat{\bm{\epsilon}}^{k}_{k+1} + J^{\epsilon}_{\omega }\delta \bm{b}_{\omega } + J^{\epsilon}_{\rho}\delta \bm{\rho}.
\end{align}
This technique is critical for enabling real-time computation of the Cerberus while doing kinematic calibration. 

\subsection{Contact-Aware Measurement Noise}
Contact preintegration can only serve as a valid measurement when the robot foot is stationary between two time steps. We reflect this fact in the measurement noise. 

Assume the robot is able to get a contact flag $c\in\{0,1\}$ indicating whether the foot is in contact (1) or not (0). The flag may come from a foot contact sensor or an estimation algorithm \cite{hwangbo2016probabilistic}. For robots without contact sensors, we leverage a standard outlier-rejection method common in Kalman Filter implementations \cite{bloesch2013slip} that fuses IMU information and the LO velocity. If the filter treats a leg as stationary according to a prior contact schedule, then velocity calculated using \eqref{eqn:lo} should agree with the current robot body velocity estimation. Otherwise, the prior contact schedule is wrong so the actual contact flag should be reverted. 

For the noise covariances in \ref{eqn:error-dyn-leg}, we let 
\begin{align}
    \sigma_{\rho} & =  c \sigma_c +  (1-c)\sigma_{nc} \ \text{and} \\
    \sigma_{v} & =  c \sigma_{0} +  (1-c)\sigma_{1},
\end{align}
which means we give the kinematic parameter and velocity measurement low uncertainty values when the foot has contact, otherwise the uncertainty is high so it does not got updated as aggressively. $\sigma_c$, $\sigma_{nc}$, $\sigma_{0}$, and $\sigma_{1}$ are all tunable hyper-parameters of the measurement model.

\section{Experiments}\label{sec:experiments}
Our C++ implementation of the Cerberus uses the factor graph optimizer and vision front end of the open source visual-inertial odometry software VINS-Fusion \cite{qin2018vins}. The IMU factor in VINS-Fusion is replaced with our proposed cost function \eqref{eqn:residual-leg}.  We set $\bm{\rho} = [l_c]$, the calf length shown in Fig. \ref{fig:frame} as it is changing during locomotion \cite{yang2022online}. We conducted experiments on sensor data collected on two quadruped robot platforms, the Unitree A1 and Go1 \cite{A1}. Both robots perform trotting using different controller implementations. The list of sensors that provide data to our state estimator is summarized in Table \ref{tab:hardware}. 

We focus on comparing the position drift percentages of a Kalman Filter (KF) \cite{bloesch2013state}, visual-inertial odometry (VINS) \cite{qin2018vins}, visual-inertia-leg odometry without kinematics calibration (VILO w/o calib), and the Cerberus (VILO with calib). The only difference between the last two is the VILO w/o calib just uses a fixed value $\bm{\rho} = [0.21m]$ while the Cerberus calibrates the kinematic parameters. 

\subsection{Indoor Experiments}
In a lab space equipped with an OptiTrack\cite{OptiTrack} motion-capture system, the robot moves on flat ground following different paths with an average speed of 0.5m/s. We record sensor data and ground-truth positions. We then run the Cerberus on a desktop computer with Intel i7-7800X 3.50GHz CPU. The processing time is 50ms per camera frame on average, which is faster than the camera sample rate (66ms). Therefore, the state estimator should run in real time. 

\begin{figure}
    \centering
    \resizebox{0.8\linewidth}{!}{%
        \input{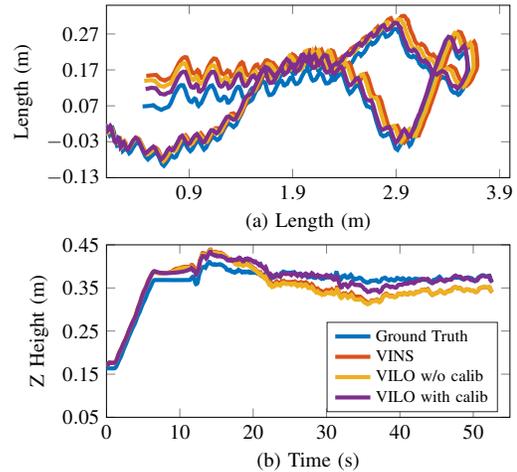}
    }
    \caption{
        Comparing with the mocap ground truth, VILO with calib has smaller drift on all directions. The final drift of the VINS trajectory (red) is 1.73\% while the drift of the VILO w/o calib trajectory (yellow) is 1.25\% and that of VILO with calib (purple) is 1.13\%. 
      }
    \label{fig:indoor_data}
\end{figure}
Figure \ref{fig:indoor_data} compares the ground truth trajectory (blue) with estimated trajectory using VINS (red), VILO w/o calib (yellow), and VILO with calib (purple) in one dataset. Table \ref{tab:outdoor_experiments} shows average performance over 10 datasets. 
\subsection{Outdoor Experiments}
The contribution of kinematics calibration to long-term position estimation is verified in outdoor experiments. Two robots collected datasets in several outdoor environments while traveling over 1.5 km with an average velocity of 0.5 $m/s$. Note that our robots move at a much faster speed than prior works (for example, \cite{kim2022step} is 0.125 $m/s$ and \cite{wisth2022vilens} is 0.25 $m/s$). In each dataset, the robot moves in a large loop and we evaluate the final position estimation drift after the robot returns to the starting point. We also note that 1\% drift is equivalent to $0.1m$ of the 10M Relative Translation Error (RTE) metric used in \cite{wisth2022vilens} and \cite{kim2022step}. Details of datasets can be found in the open-source code base.

\begin{figure}
    \centering
    \begin{minipage}[b]{.30\linewidth}
        \includegraphics[width=\linewidth]{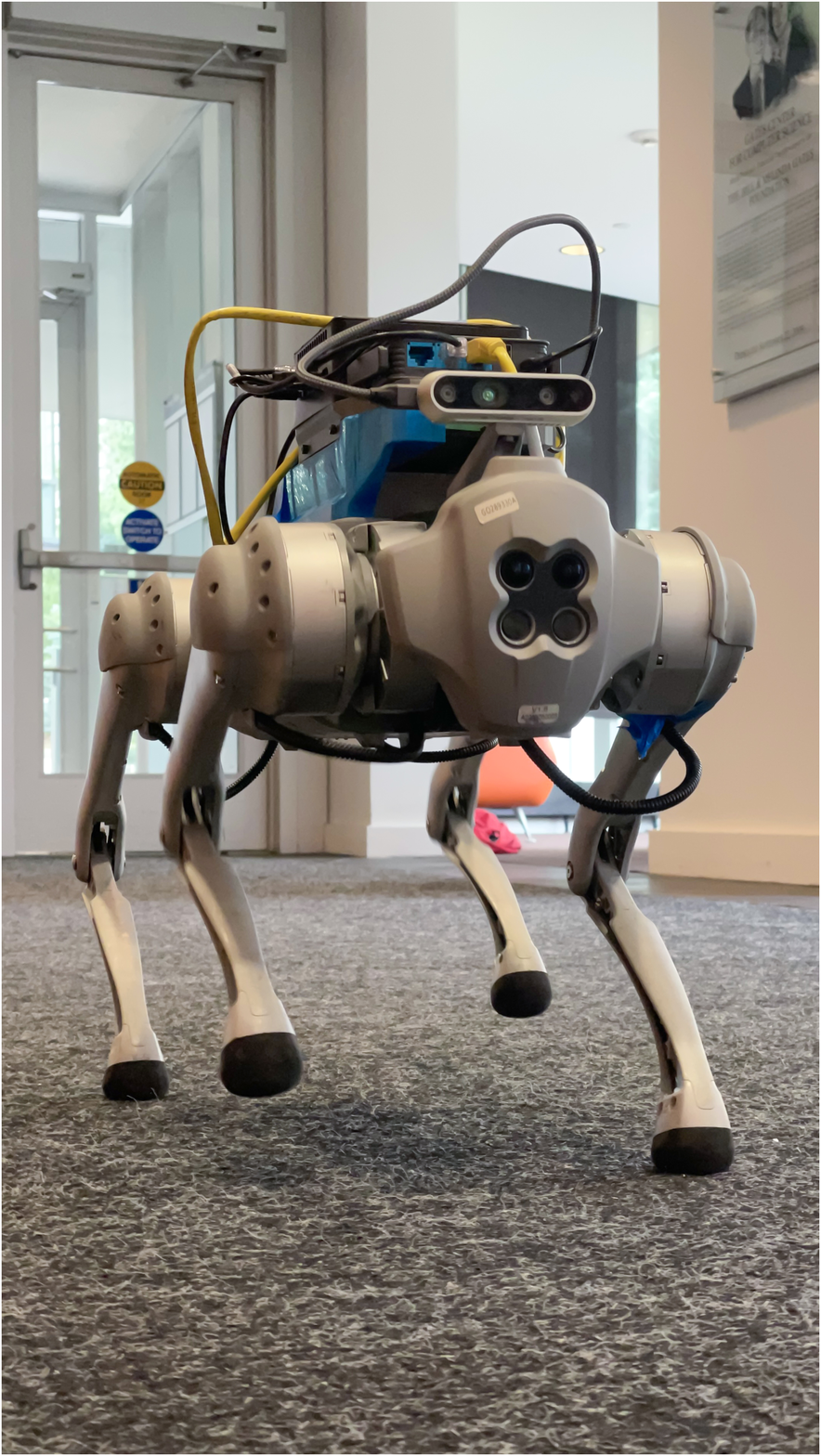}
    \end{minipage}%
    \hfill
    \begin{minipage}[b]{.685\linewidth}
        \includegraphics[width=\linewidth]{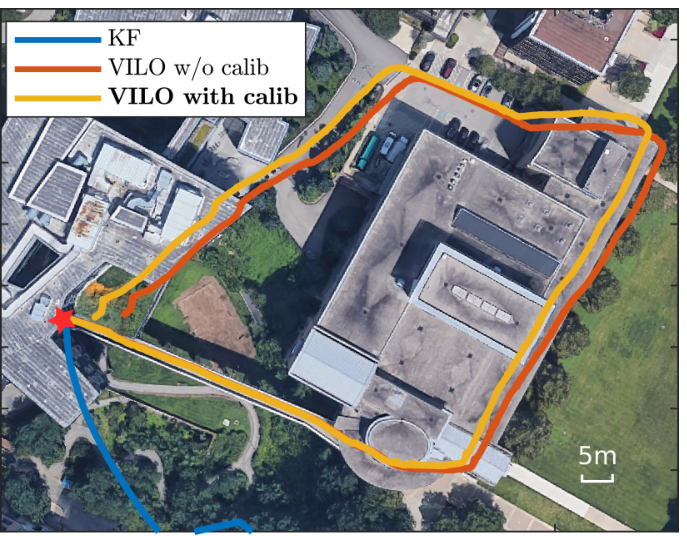}
    \end{minipage}
    
    \caption{During the recording of the ``Campus'' dataset, the Go 1 robot ran 345 m with an average speed of 1 m/s in indoor and outdoor environments. VINS fails, so no result is shown. VILO with calibration has the smallest final position drift after returning to the starting point (red star).}
    
    \label{fig:ghc-drama}
\end{figure}

\begin{figure}

    \centering
    \begin{minipage}[t]{.35\linewidth}
        \strut\vspace*{-0.7\baselineskip}\newline
        \includegraphics[width=\linewidth]{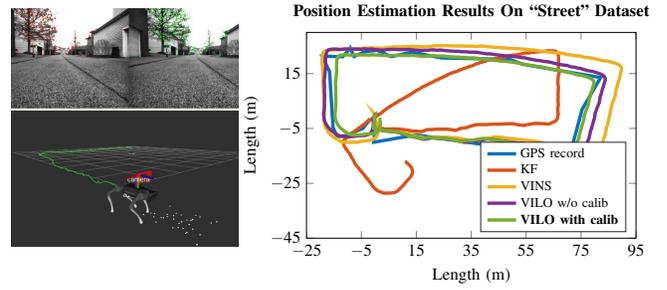}
    \end{minipage}%
    \hfill
    \begin{minipage}[t]{.65\linewidth}
    \strut\vspace*{-\baselineskip}\newline
        \resizebox{\linewidth}{!}{%
        \input{images/street_data.tex}
        }
    \end{minipage}
    
    \caption{Dataset ``Street'' algorithm run visualization and the estimation result. GPS position reference is collected using iPhone App ``Gaia GPS''. The final drift of VILO with calib is 2.22m (0.85\% after 260m travel) comparing to 1.84m of VILO w/o calib.}
    \label{fig:vilo_street_compare}
\end{figure}

Figures \ref{fig:title_pic}, \ref{fig:ghc-drama}, and \ref{fig:vilo_street_compare} compare the estimated trajectories for three datasets, ``Track'', ``Campus'', and ``Street''. Table \ref{tab:outdoor_experiments} contains quantitative analysis of drift percentage for different datasets. The ``Campus'' dataset is particularly difficult because the robot runs at over 1 $m/s$ on various indoor and outdoor terrains with different slopes. See the supplementary video for its estimation run visualization and kinematic parameter estimation result. VILO with calib outperforms all other methods across all datasets except for ``Street'', where both methods have very small drift values that have no statistically significant difference. Even though our datasets are longer and contain faster and more challenging dynamics, the Cerberus algorithm achieves $<1\%$ drift on most of them and $1.65\%$ drift on the hardest case. No prior work has achieved this level of performance.

\begin{table}[]
\small
\centering
\begin{tabular}{|c|c|c|c|c|}
\hline 
Dataset & KF & VINS & \makecell{VILO  \\ w/o calib} & \makecell{VILO \\ with calib} \\
\hline
\makecell{Indoor \\ (average 10)} & 6.53\% & 1.31\% & 1.02\% & \textbf{0.92\%} \\
\hline 
Street & $>$ 10\% & 0.89\%  & \textbf{0.70\%}  & 0.85\% \\
\hline 
Track & $>$ 10\% & 3.9\%  & 2.6\%  & \textbf{0.98\%} \\
\hline 
Campus & $>$ 10\% & break  & 3.32\%  & \textbf{1.65\%} \\
\hline

\end{tabular}
\caption{Hardware Experiment Final Drifts Comparison}
\label{tab:outdoor_experiments}
\end{table}

\subsection{Robust Estimation}
Since the Cerberus combines various sensor sources, the position estimation is robust against camera occlusion, foot slippage, and excessive body shakiness. The supplementary video contains more challenging scenarios that demonstrate the robustness of the estimator.

\section{Conclusions}\label{sec:conclude}
We have presented the Cerberus, a VILO algorithm using kinematics calibration in contact preintegration and contact outliner rejection to improve performance. Indoor and outdoor experiments on two robots have demonstrated that our state estimator outperforms many existing methods. We believe kinematics parameter error, like IMU biases, should always be modeled and calibrated to achieve precise long-term estimation for legged robots. Finally, our open-sourced Cerberus package can serve as a baseline for future work.

\appendix
\addcontentsline{toc}{section}{Appendix}
\renewcommand{\thesubsection}{\Alph{subsection}}

In \eqref{eqn:hat_vm}, the $\hat{\bm{v}}$ is an estimation. Now write a ``true'' measurement considering noisy system state and expand as
\begin{align}
\bm{v}_{m} =& -J(\bm{\phi}-\bm{n}_{\phi},\bm{\rho})(\dot{\bm{\phi}}-\bm{n}_{\dot{\phi}}) \nonumber \\
& - \skewcross{\bm{\omega}_{m}-\bm{b}_{\omega}-\bm{n}_\omega}g(\bm{\phi}-\bm{n}_{\phi},\bm{\rho}) \\
= & -J(\bm{\phi},\bm{\rho})\dot{\bm{\phi}} - \skewcross{\bm{\omega}_{m}-\bm{b}_{\omega}} g(\bm{\phi}, \bm{\rho}) \nonumber \\
& - \skewcross{g(\bm{\phi}, \bm{\rho})}\bm{n}_{\omega} +  J(\bm{\phi},\bm{\rho})\bm{n}_{\dot{\phi}} \nonumber\\
& + [(\dot{\bm{\phi}}^T \frownotimes I_3)\frac{\partial vec(J)}{\partial \bm{\phi}} + \skewcross{\bm{\omega}_{m}-\bm{b}_{\omega}} J]\bm{n}_{\phi},
\end{align}
where $vec(J)$ is a vertical stack of columns of $J$. $\frownotimes$ is the kronecker product.

According to the definition of $\bm{e}'_t$ and \eqref{eqn:pre-int-terms-epsilon}
\begin{align}
   \delta\dot{\bm{\epsilon}}^k_{t}  =  \dot{\bm{\epsilon}}^{k}_{t} - \dot{\hat{\bm{\epsilon}}}^{k}_{t}
   = A({\bm{\gamma}}^{b_k}_t)(\bm{v}_{m_t} + \bm{n}_{v}) - A(\hat{\bm{\gamma}}^{b_k}_t)  \hat{\bm{v}}_{m_t}. \label{eqn:apx_depsilon1}
\end{align}
Recall that $\bm{\rho} = \hat{\bm{\rho}} + \delta \bm{\rho}$ and $\bm{b}_\omega = \hat{\bm{b}}_\omega + \delta \bm{b}_\omega $. 
Continue expanding \eqref{eqn:apx_depsilon1} while ignoring second order delta terms \cite{qin2018vins},
\begin{align}
   &\delta\dot{\bm{\epsilon}}^k_{t}  =  \dot{\bm{\epsilon}}^{k}_{t} - \dot{\hat{\bm{\epsilon}}}^{k}_{t} \nonumber\\
       = &-A(\hat{\bm{\gamma}}^{b_k}_t)\skewcross{\hat{\bm{v}}_{m}}\delta \bm{\theta}^{k}_t 
   + \bm{\zeta}\delta \bm{b}_{\omega} + \bm{\kappa}\delta \bm{\rho} \nonumber \\
   & + \bm{\zeta}\bm{n}_\omega +  \bm{\eta}\bm{n}_{\phi} + A(\hat{\bm{\gamma}}^{b_k}_t)J\bm{n}_{\dot{\phi}} + \bm{n}_v,
\end{align}
where $\bm{\zeta} = -A(\hat{\bm{\gamma}}^{b_k}_t)\skewcross{g}$, 
$\bm{\kappa} = -A [(\dot{\bm{\phi}}^T \frownotimes I_{3})\frac{\partial vec(J)}{\partial \hat{\rho}} + \skewcross{\bm{\omega}_{m}-\hat{\bm{b}}_{\omega}}J]$ and $\bm{\eta} = A [(\dot{\bm{\phi}}^T \frownotimes I_{3})\frac{\partial vec(J)}{\partial \phi} + \skewcross{\bm{\omega}_{m}-\hat{\bm{b}}_{\omega}}J]$. 

\bibliographystyle{IEEEtran}
\bibliography{sections/references}

\begin{thebibliography}{10}
\providecommand{\url}[1]{#1}
\csname url@samestyle\endcsname
\providecommand{\newblock}{\relax}
\providecommand{\bibinfo}[2]{#2}
\providecommand{\BIBentrySTDinterwordspacing}{\spaceskip=0pt\relax}
\providecommand{\BIBentryALTinterwordstretchfactor}{4}
\providecommand{\BIBentryALTinterwordspacing}{\spaceskip=\fontdimen2\font plus
\BIBentryALTinterwordstretchfactor\fontdimen3\font minus
  \fontdimen4\font\relax}
\providecommand{\BIBforeignlanguage}[2]{{%
\expandafter\ifx\csname l@#1\endcsname\relax
\typeout{** WARNING: IEEEtran.bst: No hyphenation pattern has been}%
\typeout{** loaded for the language `#1'. Using the pattern for}%
\typeout{** the default language instead.}%
\else
\language=\csname l@#1\endcsname
\fi
#2}}
\providecommand{\BIBdecl}{\relax}
\BIBdecl

\bibitem{bloesch2013state}
M.~Bloesch, M.~Hutter, M.~A. Hoepflinger, S.~Leutenegger, C.~Gehring, C.~D.
  Remy, and R.~Siegwart, ``State estimation for legged robotsconsistent fusion
  of leg kinematics and imu,'' \emph{Robotics}, vol.~17, pp. 17--24, 2013.

\bibitem{bledt2018cheetah}
G.~Bledt, M.~J. Powell, B.~Katz, J.~Di~Carlo, P.~M. Wensing, and S.~Kim, ``Mit
  cheetah 3: Design and control of a robust, dynamic quadruped robot,'' in
  \emph{2018 IEEE/RSJ International Conference on Intelligent Robots and
  Systems (IROS)}.\hskip 1em plus 0.5em minus 0.4em\relax IEEE, 2018, pp.
  2245--2252.

\bibitem{camurri2020pronto}
M.~Camurri, M.~Ramezani, S.~Nobili, and M.~Fallon, ``Pronto: A multi-sensor
  state estimator for legged robots in real-world scenarios,'' \emph{Frontiers
  in Robotics and AI}, vol.~7, p.~68, 2020.

\bibitem{wisth2022vilens}
D.~Wisth, M.~Camurri, and M.~Fallon, ``Vilens: Visual, inertial, lidar, and leg
  odometry for all-terrain legged robots,'' \emph{IEEE Transactions on
  Robotics}, 2022.

\bibitem{bloesch2017state}
M.~Bloesch, ``State estimation for legged robots-kinematics, inertial sensing,
  and computer vision,'' Ph.D. dissertation, ETH Zurich, 2017.

\bibitem{dellaert2017factor}
F.~Dellaert, M.~Kaess \emph{et~al.}, ``Factor graphs for robot perception,''
  \emph{Foundations and Trends{\textregistered} in Robotics}, vol.~6, no. 1-2,
  pp. 1--139, 2017.

\bibitem{hartley2018hybrid}
R.~Hartley, M.~G. Jadidi, L.~Gan, J.-K. Huang, J.~W. Grizzle, and R.~M.
  Eustice, ``Hybrid contact preintegration for visual-inertial-contact state
  estimation using factor graphs,'' in \emph{International Conference on
  Intelligent Robots and Systems}, 2018, pp. 3783--3790.

\bibitem{wisth2019robust}
D.~Wisth, M.~Camurri, and M.~Fallon, ``Robust legged robot state estimation
  using factor graph optimization,'' \emph{IEEE Robotics and Automation
  Letters}, vol.~4, no.~4, pp. 4507--4514, 2019.

\bibitem{kim2022step}
Y.~Kim, B.~Yu, E.~M. Lee, J.-h. Kim, H.-w. Park, and H.~Myung, ``Step: State
  estimator for legged robots using a preintegrated foot velocity factor,''
  \emph{IEEE Robotics and Automation Letters}, vol.~7, no.~2, pp. 4456--4463,
  2022.

\bibitem{li2013high}
M.~Li and A.~I. Mourikis, ``High-precision, consistent ekf-based
  visual-inertial odometry,'' \emph{The International Journal of Robotics
  Research}, vol.~32, no.~6, pp. 690--711, 2013.

\bibitem{forster2015imu}
C.~Forster, L.~Carlone, F.~Dellaert, and D.~Scaramuzza, ``Imu preintegration on
  manifold for efficient visual-inertial maximum-a-posteriori
  estimation.''\hskip 1em plus 0.5em minus 0.4em\relax Georgia Institute of
  Technology, 2015.

\bibitem{qin2018online}
T.~Qin and S.~Shen, ``Online temporal calibration for monocular visual-inertial
  systems,'' in \emph{2018 IEEE/RSJ International Conference on Intelligent
  Robots and Systems (IROS)}.\hskip 1em plus 0.5em minus 0.4em\relax IEEE,
  2018, pp. 3662--3669.

\bibitem{bloesch2013slip}
M.~Bloesch, C.~Gehring, P.~Fankhauser, M.~Hutter, M.~A. Hoepflinger, and
  R.~Siegwart, ``State estimation for legged robots on unstable and slippery
  terrain,'' in \emph{2013 IEEE/RSJ International Conference on Intelligent
  Robots and Systems}.\hskip 1em plus 0.5em minus 0.4em\relax IEEE, 2013, pp.
  6058--6064.

\bibitem{wisth2020preintegrated}
D.~Wisth, M.~Camurri, and M.~Fallon, ``Preintegrated velocity bias estimation
  to overcome contact nonlinearities in legged robot odometry,'' in \emph{2020
  IEEE International Conference on Robotics and Automation (ICRA)}.\hskip 1em
  plus 0.5em minus 0.4em\relax IEEE, 2020, pp. 392--398.

\bibitem{yang2022online}
S.~Yang, H.~Choset, and Z.~Manchester, ``Online kinematic calibration for
  legged robots,'' \emph{IEEE Robotics and Automation Letters}, 2022.

\bibitem{ros}
\BIBentryALTinterwordspacing
{Stanford Artificial Intelligence Laboratory et al.}, ``Robotic operating
  system.'' [Online]. Available: \url{https://www.ros.org}
\BIBentrySTDinterwordspacing

\bibitem{merkel2014docker}
D.~Merkel, ``Docker: lightweight linux containers for consistent development
  and deployment,'' \emph{Linux journal}, vol. 2014, no. 239, p.~2, 2014.

\bibitem{scaramuzza2011visual}
D.~Scaramuzza and F.~Fraundorfer, ``Visual odometry [tutorial],'' \emph{IEEE
  robotics \& automation magazine}, vol.~18, no.~4, pp. 80--92, 2011.

\bibitem{hartley2003multiple}
R.~Hartley and A.~Zisserman, \emph{Multiple view geometry in computer
  vision}.\hskip 1em plus 0.5em minus 0.4em\relax Cambridge university press,
  2003.

\bibitem{sun2018robust}
K.~Sun, K.~Mohta, B.~Pfrommer, M.~Watterson, S.~Liu, Y.~Mulgaonkar, C.~J.
  Taylor, and V.~Kumar, ``Robust stereo visual inertial odometry for fast
  autonomous flight,'' \emph{IEEE Robotics and Automation Letters}, vol.~3,
  no.~2, pp. 965--972, 2018.

\bibitem{qin2018vins}
T.~Qin, P.~Li, and S.~Shen, ``Vins-mono: A robust and versatile monocular
  visual-inertial state estimator,'' \emph{IEEE Transactions on Robotics},
  vol.~34, no.~4, pp. 1004--1020, 2018.

\bibitem{hartley2020contact}
R.~Hartley, M.~Ghaffari, R.~M. Eustice, and J.~W. Grizzle, ``Contact-aided
  invariant extended kalman filtering for robot state estimation,'' \emph{The
  International Journal of Robotics Research}, vol.~39, no.~4, pp. 402--430,
  2020.

\bibitem{camurri2017probabilistic}
M.~Camurri, M.~Fallon, S.~Bazeille, A.~Radulescu, V.~Barasuol, D.~G. Caldwell,
  and C.~Semini, ``Probabilistic contact estimation and impact detection for
  state estimation of quadruped robots,'' \emph{IEEE Robotics and Automation
  Letters}, vol.~2, no.~2, pp. 1023--1030, 2017.

\bibitem{hwangbo2016probabilistic}
J.~Hwangbo, C.~D. Bellicoso, P.~Fankhauser, and M.~Hutter, ``Probabilistic foot
  contact estimation by fusing information from dynamics and
  differential/forward kinematics,'' in \emph{2016 IEEE/RSJ International
  Conference on Intelligent Robots and Systems (IROS)}.\hskip 1em plus 0.5em
  minus 0.4em\relax IEEE, 2016, pp. 3872--3878.

\bibitem{kim2021legged}
J.-H. Kim, S.~Hong, G.~Ji, S.~Jeon, J.~Hwangbo, J.-H. Oh, and H.-W. Park,
  ``Legged robot state estimation with dynamic contact event information,''
  \emph{IEEE Robotics and Automation Letters}, vol.~6, no.~4, pp. 6733--6740,
  2021.

\bibitem{jackson2021planning}
B.~E. Jackson, K.~Tracy, and Z.~Manchester, ``Planning with attitude,''
  \emph{IEEE Robotics and Automation Letters}, vol.~6, no.~3, pp. 5658--5664,
  2021.

\bibitem{murray2017mathematical}
R.~M. Murray, Z.~Li, and S.~S. Sastry, \emph{A mathematical introduction to
  robotic manipulation}.\hskip 1em plus 0.5em minus 0.4em\relax CRC press,
  2017.

\bibitem{lin2005leg}
P.-C. Lin, H.~Komsuoglu, and D.~E. Koditschek, ``A leg configuration
  measurement system for full-body pose estimates in a hexapod robot,''
  \emph{IEEE Transactions on robotics}, vol.~21, no.~3, pp. 411--422, 2005.

\bibitem{adams1956introduction}
D.~Adams, ``Introduction to inertial navigation,'' \emph{The Journal of
  Navigation}, vol.~9, no.~3, pp. 249--259, 1956.

\bibitem{usenko2016direct}
V.~Usenko, J.~Engel, J.~St{\"u}ckler, and D.~Cremers, ``Direct visual-inertial
  odometry with stereo cameras,'' in \emph{2016 IEEE International Conference
  on Robotics and Automation (ICRA)}.\hskip 1em plus 0.5em minus 0.4em\relax
  IEEE, 2016, pp. 1885--1892.

\bibitem{hartley2018legged}
R.~Hartley, J.~Mangelson, L.~Gan, M.~G. Jadidi, J.~M. Walls, R.~M. Eustice, and
  J.~W. Grizzle, ``Legged robot state-estimation through combined forward
  kinematic and preintegrated contact factors,'' in \emph{2018 IEEE
  International Conference on Robotics and Automation (ICRA)}.\hskip 1em plus
  0.5em minus 0.4em\relax IEEE, 2018, pp. 4422--4429.

\bibitem{realsense}
Intel, ``{Intel Realsense D435},''
  \url{https://www.intelrealsense.com/depth-camera-d435/}, 2022, [Online;
  accessed 10-Sep-2022].

\bibitem{A1}
Unitree, ``{A1},'' \url{https://www.unitree.com/products/a1/}, 2022, [Online;
  accessed 10-Sep-2022].

\bibitem{OptiTrack}
OptiTrack, ``{OptiTrack},'' \url{https://optitrack.com/}, 2022, [Online;
  accessed 10-Sep-2022].

\end{thebibliography}

\end{document}